\documentclass{article}
\usepackage{spconf,amsmath,epsfig}

\usepackage{makecell}

\title{Quantitative Assessment of DESIS Hyperspectral Data for Plant Biodiversity Estimation in Australia}
\name{Yiqing Guo, Karel Mokany, Cindy Ong, Peyman Moghadam, Simon Ferrier,  Shaun Levick}
\address{Commonwealth Scientific and Industrial Research Organisation, Australia}
\begin{document}
%
\maketitle
\begin{abstract}
Diversity of terrestrial plants plays a key role in maintaining a stable, healthy, and productive ecosystem. Though remote sensing has been seen as a promising and cost-effective proxy for estimating plant diversity, there is a lack of quantitative studies on how confidently plant diversity can be inferred from spaceborne hyperspectral data. In this study, we assessed the ability of hyperspectral data captured by the DLR Earth Sensing Imaging Spectrometer (DESIS) for estimating plant species richness in the Southern Tablelands and Snowy Mountains regions in southeast Australia. Spectral features were firstly extracted from DESIS spectra with principal component analysis, canonical correlation analysis, and partial least squares analysis. Then regression was conducted between the extracted features and plant species richness with ordinary least squares regression, kernel ridge regression, and Gaussian process regression. Results were assessed with the coefficient of correlation ($r$) and Root-Mean-Square Error (RMSE), based on a two-fold cross validation scheme. With the best performing model, $r$ is 0.71 and RMSE is 5.99 for the Southern Tablelands region, while $r$ is 0.62 and RMSE is 6.20 for the Snowy Mountains region. The assessment results reported in this study provide supports for future studies on understanding the relationship between spaceborne hyperspectral measurements and terrestrial plant biodiversity. 
\end{abstract}
\section{Introduction}
\label{sec:intro}

Diversity of terrestrial plants serves as an important factor in maintaining a stable, productive and healthy ecosystem. Timely and accurate biodiversity maps provide critical information for environmental conservation and restoration. Traditional methods for plant biodiversity mapping require labour-intensive and time-consuming field experiments to collect in-situ samples \cite{wang2019remote, guo2018effective}. The high cost of field experiments has led to a lack of ground samples, both in terms of completeness and representativeness. Therefore, data gaps and biases still remain as a major challenge in compiling up-to-date plant biodiversity maps of fine resolution and wide coverage \cite{kattge2020try}.

The DLR Earth Sensing Imaging Spectrometer (DESIS) \cite{krutz2019instrument} on-board the International Space Station has opened up a unique opportunity to monitor biodiversity from space. It delivers hyperspectral images with 235 spectral bands over the visible and near-infrared regions of 400 $\sim$ 1000 nm, with a spectral resolution of 2.55 nm and a spatial resolution 30 m \cite{alonso2019data}. The high resolutions in both the spectral and spatial domains make DESIS a promising data source for estimating plant biodiversity from space over large areas and in a timely manner. Previous studies have shown that plant biodiversity is linked to remotely sensed spectral measurements because of a well-founded interrelationship between plant species richness and primary productivity \cite{wang2016seasonal, grace2016integrative}. An important but relatively under-investigated question is how accurate the DESIS hyperspectral data is in estimating plant species richness.  

The original bands in hyperspectral data contain a high degree of redundancy due to spectral collinearity \cite{richards2006remote, jia1999segmented}. Dimensionality reduction can be conducted to project data from the original high-dimensional bands to a feature space of a lower dimensionality. Among popular algorithms for dimensionality reduction are Principal Component Analysis (PCA), Canonical Correlation Analysis (CCA), and Partial Least Squares (PLS). Compared with pre-defined vegetation indices such as Ratio Vegetation Index (RVI) and Normalized Difference Vegetation Index (NDVI), spectral features generated with data-driven dimensionality reduction have shown better performance in extracting useful information from hyperspectral measurements \cite{zhao2014early}. With the extracted spectral features, regression analysis can then be conducted to explore the underlying relationship between the extracted features and the target biological variables. Commonly applied linear and non-linear algorithms for regression include the Ordinary Least Squares Regression (OLSR), Kernel Ridge Regression (KRR), and Gaussian Process Regression (GPR). Statistical regression based on feature extraction has been shown effective in addressing biological problems with hyperspectral remote sensing data \cite{hacker2020retrieving, zhao2014early}.

In this study, we assessed the ability of DESIS hyperspectral data in plant biodiversity mapping in two regions in southeast Australia, namely the Southern  Tablelands  and  the Snowy  Mountains. Spectral features were firstly extracted from the original DESIS spectra with dimensionality reduction methods of PCA, CCA, and PLS. Then the extracted features were correlated to ground measured plant species richness via linear and non-linear regression algorithms including OLSR, KRR, and GPR, with the aim to estimate plant species richness from DESIS hyperspectal data. Accuracy assessment was conducted with a two-fold cross validation scheme, with results being reported for both regions.

\section{Materials}
\label{sec:materials}

\subsection{Ground Truth Species Richness}
\label{ssec:ground}

\begin{figure}[htbp]
\centering
\includegraphics[width=8.5cm]{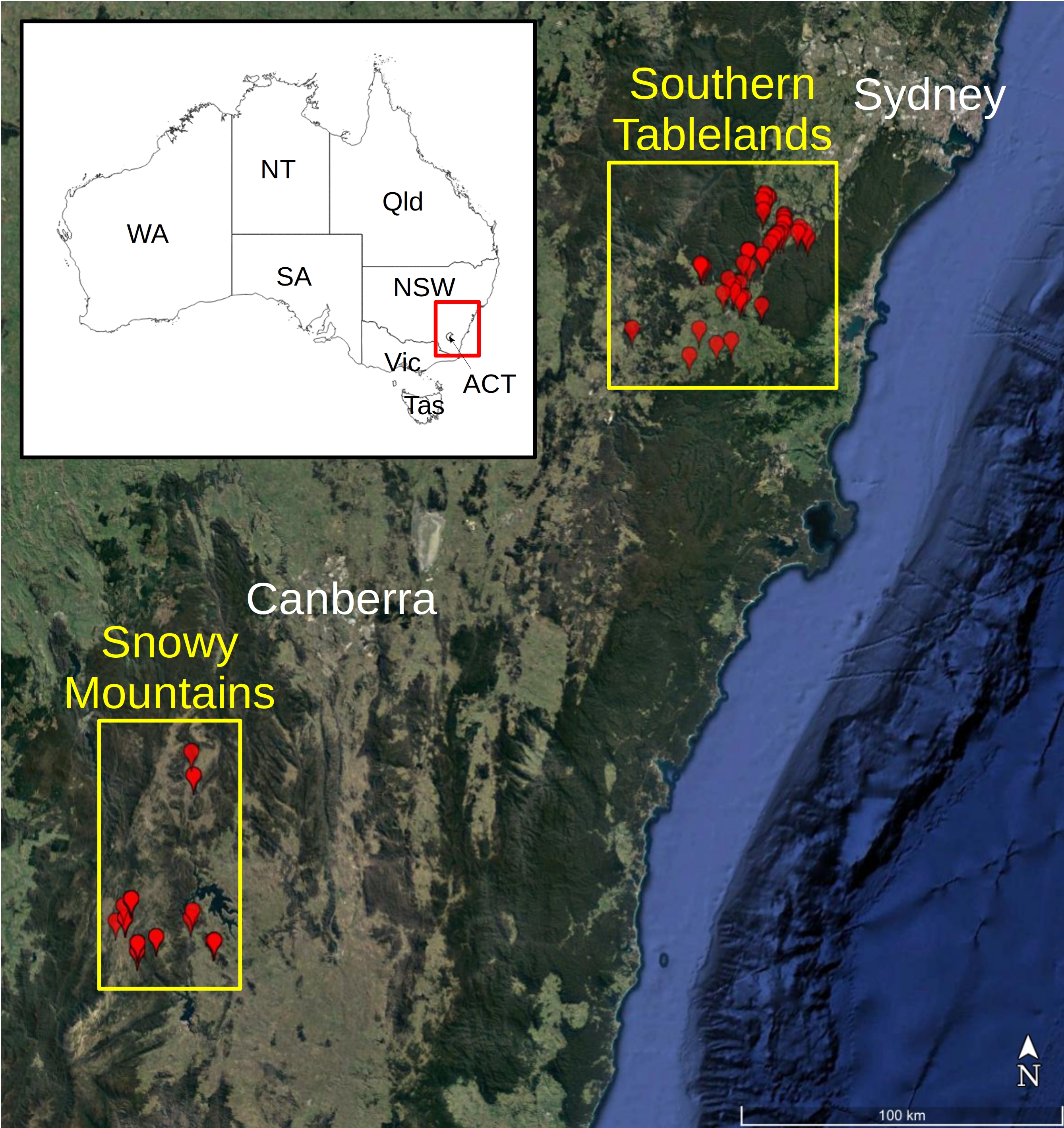}
\caption{\label{fig:location}Locations of in-situ plant species richness samples collected in field experiments.}
\end{figure}

This study focused on two regions in southeast New South Wales (NSW), Australia, namely the Southern Tablelands and the Snowy Mountains, as shown in Fig. \ref{fig:location}. The Southern Tablelands region is located to the southwest of Sydney, characterised by flat plains with a large part of the lands being transformed into pastures for grazing purposes. The Snowy Mountains region is located to the southwest of Canberra, encompassing the highest mountain ranges of the Australian Alps, with a variety of plant species including some of the most threatened in Australia. 

Field surveys were conducted to collect species richness samples in 2016 and 2017 for the two regions. For each surveying location, the size of sampling plot is 400 m$^2$. A significant event of bushfires happened during the 2019--2020 summer, with some of the sampling points situating within the affected areas. These bushfire-affected samples were excluded from the data set, based on the National Indicative Aggregated Fire Extent Datasets (NIAFED) provided by the Australian Government Department of Agriculture, Water and the Environment. After the exclusion, a total of 35 and 23 samples were used in this study for analysis for the Southern Tablelands and Snowy Mountains regions, respectively. The locations of these samples are shown in Fig. \ref{fig:location}, and their associated information is summarised in Table \ref{tab:samples}.

\begin{table*}[htbp]
\centering
\caption{\label{tab:samples} Information summary of the plant species richness samples collected in field experiments.}
\renewcommand{\arraystretch}{2}
\begin{tabular}{ |p{3cm}|p{2.4cm}|p{4cm}|p{4.5cm}|p{1.6cm}| } 
 \hline
 Region & No. of Samples & Geo-extent & Sampling Time & Plot Size\\
 \hline
 Southern Tablelands & 35 & \makecell[l]{34°12'26"--34°39'07"S \\ 150°05'57"--150°40'51"E} & Feb 19, 2017 $\sim$ Dec 07, 2017 & 400 m$^2$ \\ \hline
 Snowy Mountains & 23 & \makecell[l]{35°43'58"--36°16'30"S \\ 148°23'16"--148°39'02"E} & Feb 24, 2016 $\sim$ Dec 13, 2017 & 400 m$^2$ \\ \hline
\end{tabular}
\end{table*}

\subsection{Hyperspectral DESIS Data}
\label{ssec:hyper}

DESIS spectra intersecting with locations of the species richness samples were downloaded in surface reflectance values via CSIRO's Earth Analytics and Science Innovation (EASI) platform. These DESIS spectra were measured in January 2020. In order to moderate the random noise present in the original spectra, the spectral resolution were down-sampled into 10 nm with the assumption of a Gaussian-shaped spectral response function. The atmospherically affected bands of 759, 769, 933.4, 943.4, and 953.2 nm, and the low quality bands of 402.8, 410.3, and 999.5 nm at the left and right ends of the spectrum were removed. A total of 52 bands are retained after the removal. Spectra that are labelled as clouds are discarded. Due to the high tilting capacity of the MUSES and DESIS system (--45° to +5° across-track, --40° to +40° along-track), the spectra were mean-normalised with each spectrum being divided by its mean value over all bands, in order to minimise the bidirectional reflectance effect caused by the varied solar and sensing directions.

\section{Methods}
\label{sec:method}

\subsection{Dimensionality Reduction for Feature Extraction}
\label{ssec:feature}

The DESIS hyperspectral data provide rich information in the abundant and spectrally continuous bands, but these bands are highly collinear. To circumvent the collinearity problem, dimensionality reduction is conducted in this study. Three methods for dimensionality reduction are employed and compared, namely the Principal Component Analysis (PCA), Canonical Correlation Analysis (CCA), and Partial Least Squares (PLS). These methods aim to find a linear transformation to project DESIS spectra from the original space of $n$ spectral bands to a new space of a reduced dimensionality defined by $k$ uncorrelated components, with $k$ being smaller than $n$. 

The number of components, $k$, is a hyperparameter that needs to be pre-set. A trade-off needs to be made when selecting $k$, as a large $k$ provides more information from the original data but more redundancy is retained as well. In this study, grid search is used with $k$ as the value that achieves the best performance under the cross validation scheme as described later in Subsection \ref{ssec:accuracy}.

The PCA is an unsupervised algorithm that finds components as the ones that explain the maximum variance in the spectral data, disregarding the target values of species richness. In contrast, PLS and CCA are supervised with both spectral data and species richness values being taken into account in the computation of components. The difference between PLS and CCA is that PLS seeks to maximise the covariance between computed components and species richness values, while CCA aims to maximise the correlation between the two.

\subsection{Estimation of Plant Species Richness with Regression}
\label{ssec:plant}

After reducing the dimensionality of spectral data, the extracted components are used as features to establish correlation with plant species richness, with the aim to estimate plant species richness from DESIS data. Three commonly used regression methods are employed, including the Ordinary Least Squares Regression (OLSR), Kernel Ridge Regression (KRR), and Gaussian Process Regression (GPR). A kernel function $k(x_1, x_2)$ needs to be set for KRR and GPR. Here we use a combination of a dot-product kernel $k_{d}(x_1, x_2)$, a radial-basis function kernel $k_{r}(x_1, x_2)$, and a white kernel $k_{w}(x_1, x_2)$, as follows:
\begin{equation}
\begin{split}
k(x_1, x_2) & = k_{d}(x_1, x_2) + k_{r}(x_1, x_2) + k_{w}(x_1, x_2), \\
k_{d}(x_1, x_2) & = x_1 \cdot x_2 + \sigma^2, \\
k_{r}(x_1, x_2) &= {\rm exp}(-\frac{\left \| x_1 - x_2 \right \|_2^2}{2l^2}), \\
k_{w}(x_1, x_2) &= \delta \quad {\rm if} \; \; x_1 = x_2 \; \; {\rm else} \; \; 0,
\end{split}
\end{equation}
where $\sigma$, $l$ and $\delta$ are hyperparameters that need to be selected with grid search. The dot-product and radial-basis function kernels account for the linearity and non-linearity of the data, respectively, while the white kernel explains the noise in the data. The OLSR is a linear algorithm. By contrast, with the kernel function being specified, KRR and GPR are capable to build non-linear regression models.

\subsection{Accuracy Assessment with Cross Validation}
\label{ssec:accuracy}

A two-fold validation scheme is used in this study to assess the modelling accuracy. For each region, the whole data set is randomly partitioned into two subsets (Subsets I and II), with Subset I for training and Subset II for validation (Round I), followed by Subset II for training and Subset I for validation (Round II). This procedure is repeated 10 times with the data being partitioned differently each time. The coefficient of correlation ($r$) and Root-Mean-Square Error (RMSE) are calculated to evaluate the performance of the models.

\section{Results and Discussion}
\label{sec:results}

\subsection{Results for the Southern Tablelands Region}
\label{ssec:southern}

Different combinations of dimensionality reduction and regression methods are assessed in this study for estimating plant biodiversity with DESIS hyperspectral data. The assessment results for the Southern Tablelands region are shown in Table \ref{tab:southern_tablelands}. The best result is achieved with a combination of PLS for dimensionality reduction and GPR for regression, with $r$ being 0.71 and RMSE being 5.99. It is seen that results with PCA and PLS as the dimensionality reduction method perform better than CCA. The PCA- and PLS-based methods achieve $0.67\sim 0.70$ and $0.67\sim 0.71$ for $r$ and $6.07\sim 6.22$ and $5.99\sim 6.15$ for RMSE, respectively, better than the CCA-based methods with $0.59\sim 0.60$ for $r$ and $7.21 \sim 7.39$ for RMSE. When given a dimensionality reduction method, GPR and KRR tend to perform better OLSR for regression. The reason could be that GPR and KRR are able to account for non-linearity in the data in contrast to OLSR which is a linear algorithm.

\begin{table}[htbp]
\centering
\caption{\label{tab:southern_tablelands}Coefficient of correlation ($r$) and Root-Mean-Square Error (RMSE) between model predicted and ground truth plant species richness for the Southern Tablelands region.}
\renewcommand{\arraystretch}{1.1}
\begin{tabular}{|p{2cm}|p{1.8cm}|p{1.5cm}|p{1.5cm}|}
\cline{1-4}
Dimensionality Reduction & Regression   & $r$             & RMSE          \\ \cline{1-4}
PCA                      & OLSR       & 0.67          & 6.07          \\
PCA                      & GPR          & 0.70          & 6.22          \\
PCA                      & KRR          & 0.68          & 6.10          \\
CCA                      & OLSR       & 0.59          & 7.39          \\
CCA                      & GPR          & 0.60          & 7.29          \\
CCA                      & KRR          & 0.60          & 7.21          \\
PLS                      & OLSR       & 0.67          & 6.03          \\
\textbf{PLS}             & \textbf{GPR} & \textbf{0.71} & \textbf{5.99} \\
PLS                      & KRR          & 0.67          & 6.15          \\
\cline{1-4}
\end{tabular}
\end{table}

\subsection{Results for the Snowy Mountains Region}
\label{ssec:snowy}

The assessment results for the Snowy Mountains region in show in Table \ref{tab:snowy_mountains}. The best $r$ and RMSE are 0.62 and 6.20, respectively, achieved with a combination of PLS for dimensionality reduction and KRR for regression. By comparing to the results of the Southern Tablelands in Table \ref{tab:southern_tablelands}, it can be seen that generally the $r$ values are lower and RMSE are higher for the Snowy Mountains region. The following observations for the Southern Tablelands region (Table \ref{tab:southern_tablelands}) are also seen for the Snowy Mountains region (Table \ref{tab:snowy_mountains}): (1) with CCA for dimensionality reduction, the results are generally worse than those with PCA and PLS, and (2) the non-linear regression algorithms GPR and KRR outperform the linear OLSR. 

\begin{table}[htbp]
\centering
\caption{\label{tab:snowy_mountains}Coefficient of correlation ($r$) and Root-Mean-Square Error (RMSE) between model predicted and ground truth plant species richness for the Snowy Mountains region.}
\renewcommand{\arraystretch}{1.1}
\begin{tabular}{|p{2cm}|p{1.8cm}|p{1.5cm}|p{1.5cm}|}
\cline{1-4}
Dimensionality Reduction & Regression   & $r$             & RMSE          \\ \cline{1-4}
PCA                      & OLSR       & 0.50          & 7.24          \\
PCA                      & GPR        & 0.51          & 7.11          \\
PCA                      & KRR        & 0.54          & 6.43          \\
CCA                      & OLSR       & 0.22          & 9.83          \\
CCA                      & GPR        & 0.24          & 9.74          \\
CCA                      & KRR        & 0.24          & 9.78          \\
PLS                      & OLSR       & 0.61          & 7.11          \\
PLS                      & GPR        & 0.60          & 6.28          \\
\textbf{PLS}             & \textbf{KRR} & \textbf{0.62} & \textbf{6.20} \\
\cline{1-4}
\end{tabular}
\end{table}

\section{Conclusion}
\label{sec:conclusion}
The ability of hyperspectral data captured by the DLR Earth Sensing Imaging Spectrometer (DESIS) in estimating plant species richness was assessed in this study, with the Southern Tablelands and Snowy  Mountains regions in southeast Australia being focused for the experiments. Spectral features were firstly extracted from DESIS spectra with principal component analysis, canonical correlation analysis, and partial least squares analysis. Then regression was conducted between the extracted features and plant species richness with ordinary least squares regression, kernel ridge regression, and Gaussian process regression. The results were assessed with the coefficient of correlation ($r$) and Root-Mean-Square Error (RMSE), based on a two-fold cross validation scheme. With the best performing model, $r$ is 0.71 and RMSE is 5.99 for the Southern Tablelands region, while $r$ is 0.62 and RMSE is 6.20 for the Snowy Mountains region. The assessment results reported in this study provide supports for future studies on understanding the relationship between spaceborne hyperspectral measurements and terrestrial plant biodiversity. Future work would be on including remote sensing data from more sources to increase the estimation accuracy, and extending the current approach to larger areas.
\bibliographystyle{IEEEbib}
\bibliography{igarss}

\begin{thebibliography}{10}

\bibitem{wang2019remote}
Ran Wang and John~A Gamon,
\newblock ``Remote sensing of terrestrial plant biodiversity,''
\newblock {\em Remote Sensing of Environment}, vol. 231, pp. 111218, 2019.

\bibitem{guo2018effective}
Yiqing Guo, Xiuping Jia, and David Paull,
\newblock ``Effective sequential classifier training for {SVM}-based
  multitemporal remote sensing image classification,''
\newblock {\em IEEE Transactions on Image Processing}, vol. 27, no. 6, pp.
  3036--3048, 2018.

\bibitem{kattge2020try}
Jens Kattge, Gerhard B{\"o}nisch, Sandra D{\'\i}az, et~al.,
\newblock ``Try plant trait database--enhanced coverage and open access,''
\newblock {\em Global Change Biology}, vol. 26, no. 1, pp. 119--188, 2020.

\bibitem{krutz2019instrument}
David Krutz, Rupert M{\"u}ller, Uwe Knodt, et~al.,
\newblock ``The instrument design of the {DLR} {E}arth sensing imaging
  spectrometer ({DESIS}),''
\newblock {\em Sensors}, vol. 19, no. 7, pp. 1622, 2019.

\bibitem{alonso2019data}
Kevin Alonso, Martin Bachmann, Kara Burch, et~al.,
\newblock ``Data products, quality and validation of the {DLR} {E}arth sensing
  imaging spectrometer ({DESIS}),''
\newblock {\em Sensors}, vol. 19, no. 20, pp. 4471, 2019.

\bibitem{wang2016seasonal}
Ran Wang, John~A Gamon, Rebecca~A Montgomery, et~al.,
\newblock ``Seasonal variation in the {NDVI}--species richness relationship in
  a prairie grassland experiment ({C}edar {C}reek),''
\newblock {\em Remote Sensing}, vol. 8, no. 2, pp. 128, 2016.

\bibitem{grace2016integrative}
James~B Grace, T~Michael Anderson, Eric~W Seabloom, et~al.,
\newblock ``Integrative modelling reveals mechanisms linking productivity and
  plant species richness,''
\newblock {\em Nature}, vol. 529, no. 7586, pp. 390--393, 2016.

\bibitem{richards2006remote}
John~A Richards and Xiuping Jia,
\newblock {\em Remote Sensing Digital Image Analysis [4th Edition]},
\newblock Springer, 2006.

\bibitem{jia1999segmented}
Xiuping Jia and John~A Richards,
\newblock ``Segmented principal components transformation for efficient
  hyperspectral remote-sensing image display and classification,''
\newblock {\em IEEE Transactions on Geoscience and Remote Sensing}, vol. 37,
  no. 1, pp. 538--542, 1999.

\bibitem{zhao2014early}
Feng Zhao, Yanbo Huang, Yiqing Guo, et~al.,
\newblock ``Early detection of crop injury from glyphosate on soybean and
  cotton using plant leaf hyperspectral data,''
\newblock {\em Remote Sensing}, vol. 6, no. 2, pp. 1538--1563, 2014.

\bibitem{hacker2020retrieving}
Paul~W Hacker, Nicholas~C Coops, Philip~A Townsend, et~al.,
\newblock ``Retrieving foliar traits of {Q}uercus garryana var. garryana across
  a modified landscape using leaf spectroscopy and {L}i{DAR},''
\newblock {\em Remote Sensing}, vol. 12, no. 1, pp. 26, 2020.

\end{thebibliography}

\end{document}